\documentclass{bmvc2k}

\title{Semi-supervised Skin Lesion Segmentation \\via  Transformation Consistent Self-ensembling Model}

\addauthor{Xiaomeng Li}{xmli@cse.cuhk.edu.hk}{1}
\addauthor{Lequan Yu}{lqyu@cse.cuhk.edu.hk}{1}
\addauthor{Hao Chen}{hchen@cse.cuhk.edu.hk}{12}
\addauthor{Chi-Wing Fu}{cwfu@cse.cuhk.edu.hk}{1}
\addauthor{Pheng-Ann Heng}{pheng@cse.cuhk.edu.hk}{1}

\addinstitution{
	Department of Computer Science and Engineering\\
	The Chinese University of Hong Kong\\
	Hong Kong
}
\addinstitution{
	Imsight Medical Technology, Inc.\\
}

\runninghead{Li et al.}{Semi-supervised Skin Lesion Segmentation}

\def\eg{\emph{e.g}\bmvaOneDot}

\def\etal{\emph{et al}\bmvaOneDot}

\usepackage{times}
\usepackage{threeparttable}
\usepackage{xcolor}
\usepackage{soul}
\usepackage[utf8]{inputenc}
\usepackage[small]{caption}
\usepackage{graphicx}
\usepackage{algorithm}  
\usepackage{amsfonts}
\usepackage{algpseudocode}  
\usepackage{multirow}
\usepackage{amsmath,booktabs}
\usepackage{siunitx}
\usepackage{subfigure}
\usepackage{capt-of}
\usepackage{booktabs}
\usepackage{varwidth}

\newsavebox\tmpbox
\newcolumntype{L}[1]{>{\raggedright\let\newline\\\arraybackslash\hspace{0pt}}m{#1}}
\newcolumntype{C}[1]{>{\centering\let\newline\\\arraybackslash\hspace{0pt}}m{#1}}
\newcolumntype{R}[1]{>{\raggedleft\let\newline\\\arraybackslash\hspace{0pt}}m{#1}}

\newcommand{\TODO}[1]{{\color{red}{#1}}}

\newcommand{\para}[1]{\vspace{.05in}\noindent\textbf{#1}}

\def\ie{\emph{i.e.}}
\def\eg{\emph{e.g.}}
\def\etal{{\em et al.}}

\title{Semi-supervised Skin Lesion Segmentation \\via  Transformation Consistent Self-ensembling Model}

\begin{document}
\maketitle
\begin{abstract}
Automatic skin lesion segmentation on dermoscopic images is an essential component in computer-aided diagnosis of melanoma. 
Recently, many fully supervised deep learning based methods have been proposed for automatic skin lesion segmentation.
However, these approaches require massive pixel-wise annotation from experienced dermatologists, which is very costly and time-consuming.
In this paper, we present a novel semi-supervised method for skin lesion segmentation, where the network is optimized by the weighted combination of a common supervised loss for labeled inputs only and a regularization loss for both labeled and unlabeled data.
To utilize the unlabeled data, our method encourages the consistent predictions of the network-in-training for the same input under different regularizations.
Aiming for the semi-supervised segmentation problem, we enhance the effect of regularization for pixel-level predictions by introducing a transformation, including rotation and flipping, consistent scheme in our self-ensembling model.
With only 300 labeled training samples, our method sets a new record on the benchmark of the International Skin Imaging Collaboration (ISIC) 2017 skin lesion segmentation challenge.
Such a result clearly surpasses fully-supervised state-of-the-arts that are trained with 2000 labeled data. 
\end{abstract}
\section{Introduction}
\begin{figure}[t]
	\centering
	\includegraphics[width=0.95\linewidth]{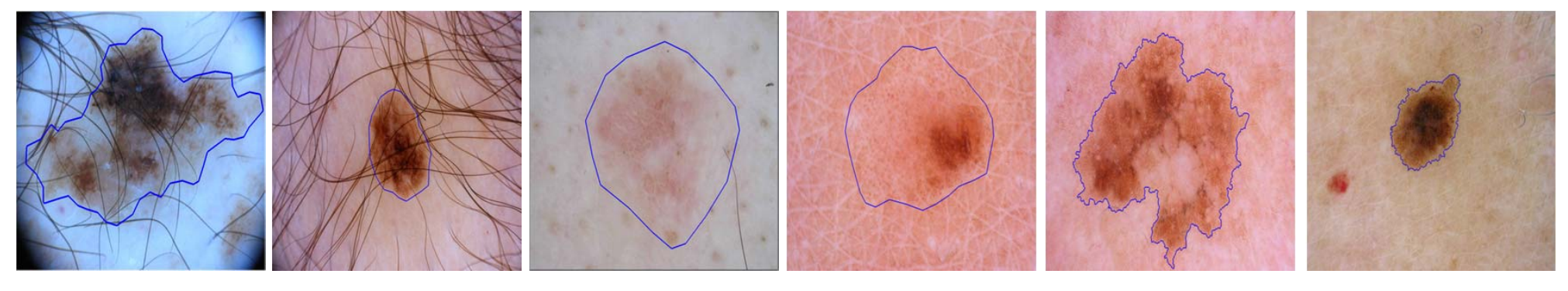}
	\caption{Skin lesion cases with artifacts (the left two); examples of ambiguous (the middle two) and clear-cut (the right two) labels.}
	\label{fig:intro}\centering
	\vspace{-0.3cm}
\end{figure}

Skin cancer is currently one of the fastest growing cancers worldwide, and melanoma is the most deadly form of skin cancer, leading to an estimated 9,730 deaths in the United States in 2017~\cite{CAAC:CAAC21387}.
To improve the diagnostic performance of melanoma,
dermoscopy has been proposed as a noninvasive imaging technique to enhance the visual effect of pigmented skin lesions.
However, recognizing malignant melanoma by visual interpretation alone is time-consuming and error-prone to inter- and intra-observer variabilities. To assist dermatologists in the diagnosis, an automatic melanoma segmentation method is highly demanded in the clinical practice.

Automatic melanoma segmentation is a very challenging task due to large variations in lesion size, location, shape and color over different patients and the presence of artifacts such as hairs and veins; see Figure~\ref{fig:intro}. 
Traditional segmentation methods are mainly based on clustering, intensity thresholding, region growing, and deformable models. 
These methods, however, rely on hand-crafted features, and have limited feature representation capability.
Recently, convolutional neural networks (CNNs) have been widely used and achieved remarkable success in a variety of vision recognition tasks.
Many researchers advanced the skin lesion segmentation and showed decent results~\cite{yuan2017improving,berseth2017isic,codella2017skin,li2018deeply}. 
For example, Yuan \etal ~\cite{yuan2017improving} proposed a deep convolutional neural network (DCNN), trained it with multiple color spaces, and achieved the best performance in the ISIC 2017 skin lesion segmentation challenge.

All the above methods, however, are based on fully supervised learning, which requires a large amount of annotated images to train the network for accuracy and robustness.
Such pixel-level annotation is laborious and difficult to obtain, especially for melanoma in the dermoscopic images, which rely heavily on experienced dermatologists.
Moreover, the limited amount of labeled data with pixel-wise annotations also restricts the performance of deep networks.
Lastly, there exists some cases that display ambiguous melanocytic or borderline  features of melanoma. 
These cases are inherently difficult to have an accurate annotation from the dermoscopic diagnosis~\cite{scolyer2010histologically}; see again Figure~\ref{fig:intro}. 
Previous supervised learning based methods do not have specific schemes to deal with these ambiguous annotations, which may degrade the performance on those dermoscopic images with clear-cut lesions. 
To alleviate the above issues, we address the skin lesion segmentation problem via semi-supervised learning, which leverages both a limited amount of labeled and an arbitrary amount of unlabeled data. 
As a by-product, our semi-supervised method is robust and has a potential to be tolerant to ambiguous labels; see experiments in Section 4.2.
There are some semi-supervised approaches for dermoscopy images and other medical image processing~\cite{masood2015self,jaisakthi2017automatic,gu2017semi,bai2017semi}.
However, they  either suffer from limited representation capacity of hand-crafted features or may easily get into local minimum.

In this paper, we present a novel semi-supervised learning method for skin lesion segmentation.
The whole framework is trained with a weighted combination of the supervised loss and the unsupervised loss.
To utilize the unlabeled data, our self-ensembling method encourages the consistent prediction of the network for the same input data under different regularizations (\eg, randomized Gaussian noise, network dropout and randomized data transformation). 
In particular, we design our method to account for the challenging semi-supervised segmentation task, in which pixel-level classification is required to be predicted.
We observe that in the segmentation problem, if one transforms (\eg, rotate) the input image, the expected prediction should be transformed in the same manner. 
Actually, when the inputs of CNNs are rotated, the corresponding network predictions would not rotated in the same way~\cite{worrall2017harmonic}. 
In this regard, we take advantages of this property by introducing a transformation (\ie, rotation, flipping) consistent scheme at the input and output space of our network. 
Specifically, we design the unsupervised/regularization loss by minimizing the differences between the network predictions under different transformations of the same input.

In summary, our work has the following achievements: 
\begin{itemize}
	\item We present a novel semi-supervised learning method for the practical biomedical image segmentation problem by taking advantage of a large amount of unlabeled data, which largely reduces annotation efforts for the dermatologists.
		
	\item To better utilize the unlabeled data for segmentation tasks, we propose a transformation consistent scheme in self-ensembling model and demonstrate the effectiveness for semi-supervised learning.
	

	\item 
	We establish a new record with only 300 labeled data on the benchmark of ISIC 2017 skin lesion segmentation challenge, which excels the state-of-the-arts that are based on fully supervised learning with 2000 labeled data.

\end{itemize}

\vspace{-0.5cm}
\section{Related Work}
\vspace{-0.3cm}
\para{Skin lesion segmentation.} \
Early approaches on skin lesion segmentation mainly focused on thresholding~\cite{emre2013lesion}, iterative/statistical region merging~\cite{iyatomi2006quantitative} and machine learning related methods~\cite{he2012automatic,sadri2013segmentation}.
Recently, many researchers employed deep learning based methods for skin lesion segmentation.
For example, Yu \emph{et al.}~\cite{yu2017automated} explored the network depth property and developed a deep residual network with more than 50 layers for automatic skin lesion segmentation, where several residual blocks were stacked together to increase the network representative capability.
Bi \emph{et al.}~\cite{bi2017dermoscopic} proposed a multi-stage approach to segment skin lesion by combining the results from multiple cascaded fully convolutional networks.
Yuan \emph{et al.}~\cite{yuan2017automatic} proposed a 19-layer deep convolutional neural network and trained it in an end-to-end manner for skin lesion segmentation. 
However, these approaches are based on fully supervised learning, requiring massive pixel-wise annotations from experienced dermatologists to create a training dataset.

\para{Transformation equivariant representation.} \
There is a body of related literature on equivariance representations, where 
the transformation equivariance is encoded to the network to explore the network equivariance property~\cite{cohen2016group, pmlr-v48-dieleman16, worrall2017harmonic}. 
For example, \citet{cohen2016group} proposed group equivariant neural network to improve the network generalization, where equivariance to \ang{90}-rotations and dihedral flips is encoded by copying the transformed filters at different rotation-flip combinations.  
Concurrently, ~\citet{pmlr-v48-dieleman16} designed four different equivariance to preserve feature map transformations by rotating feature maps instead of filters.
Recently, \citet{worrall2017harmonic} restricted  the filters to circular harmonics to achieve continuous \ang{360}-rotations equivariance.
However, these works aim to encode equivariance to the network to improve the generalization capability of the network, while our method targets to better utilize the unlabeled data in the semi-supervised learning.

\para{Semi-supervised segmentation for medical images.} \
Semi-supervised approaches have been applied in various medical imaging tasks.
Portela \emph{et al.}~\cite{portela2014semi} employed Gaussian Mixture Model (GMM) to automatically segment brain MR images.
For retinal vessel segmentation, You \emph{et al.}~\cite{you2011segmentation} combined radial projection and semi-supervised learning.
Gu \emph{et al.}~\cite{gu2017semi} proposed a semi-supervised method to segment vessel by constructing forest oriented super pixels.
While Sedai \emph{et al.}~\cite{sedai2017semi} introduced a variational autoencoder for optic cup segmentation in retinal fundus images.
There are also some  semi-supervised works for diagnosing in dermoscopy images~\cite{masood2015self,jaisakthi2017automatic}.  
For example, Jaisakthi \etal~\cite{jaisakthi2017automatic} proposed a semi-supervised skin lesion segmentation method using K-means clustering  on color features. 
However, these methods are based on hand-crafted features, which suffer from limited representation capacity.
Recently, as the surprising results achieved by CNNs in the supervised learning, semi-supervised approaches with CNNs started to attract attentions in the medical imaging field. For example, 
Bai \emph{et al.}~\cite{bai2017semi} proposed a semi-supervised fully convolutional neural network (FCN) to segment cardiac from MR images, where the network parameters and the segmentations for unlabeled data were alternatively updated. However, this method was trained offline and may easily get into local minimum.
\begin{figure*}[!t]
	\centering
	\includegraphics[width=0.98\linewidth]{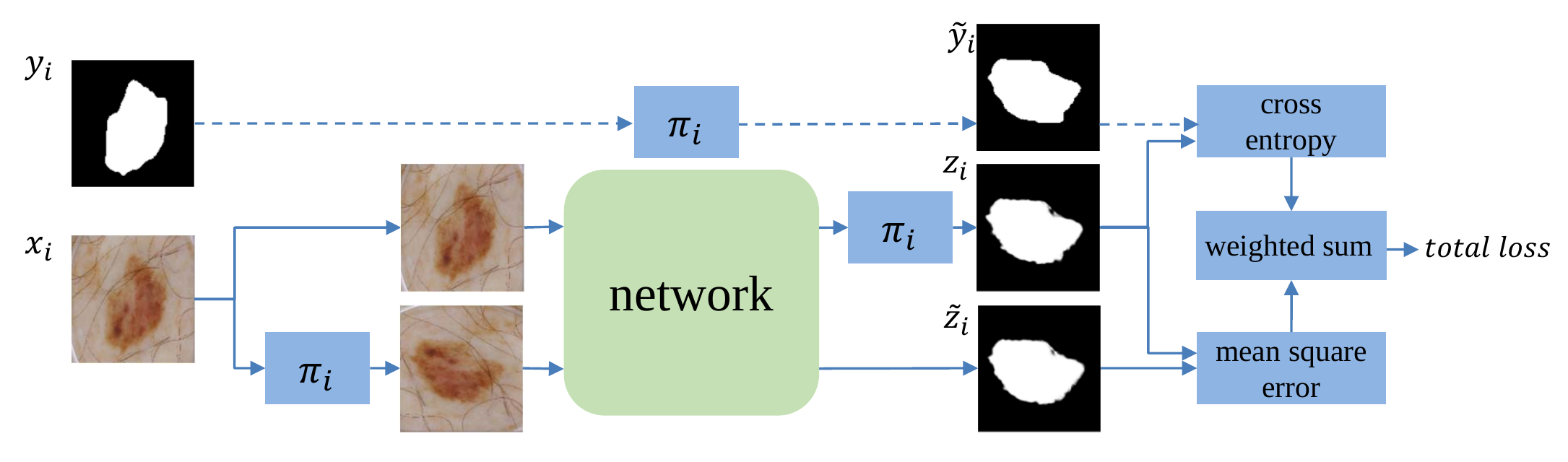}
	\caption{The pipeline of our proposed transformation consistent self-ensembling model for semi-supervised skin lesion segmentation. 
		The total loss is weighted combination of the cross-entropy loss on labeled data, and mean square error loss on both labeled and unlabeled data. 
		The model encourages the network to be transformation consistent to utilize the unlabeled data. 
		Note that $\pi_i$ remains the same in each training pass but are changed at different passes.}
	\label{fig:pipeline}\centering
	
\end{figure*}

\section{Method}

\subsection{Overview} 
We first describe the overview of our semi-supervised segmentation method. 
The training set consists $N$ inputs in total, including $M$ labeled inputs and $N-M$ unlabeled inputs. Let $\mathcal{L} = \left \{ (x_i, y_i) \right \}_{i=1}^{M}$ be the labeled set and $\mathcal{U} = \left \{x_i \right \}_{i=M+1}^{N}$ be the unlabeled set, where $x_i \in \mathbb{R}^{H \times W \times 3}$ is the input image and $y_i \in \{0,1\}^{H \times W}$ is the ground-truth label. 
Our proposed semi-supervised segmentation method can be formulated to learn the network parameters $\theta$ by optimizing:
\begin{equation}
\min_{\theta} \sum_{i=1}^{M} l (f(x_i; \theta), y_i) + \lambda R(\theta, \mathcal{L}, \mathcal{U}),
\label{eq:semi}
\end{equation}
where $f(\cdot)$ represents the network mapping, $l$ is the supervised loss function and $R$ is the regularization (unsupervised) loss.
The first component is designed for supervised training, optimized by the cross-entropy loss and used for evaluating the correctness of network output on labeled inputs only.
While the second regularization component is designed for unsupervised training by regularizing the network output on both labeled and unlabeled inputs. 
$\lambda$ is a weighting factor that controls how strong the regularization is. 

Self-ensembling methods~\cite{sajjadi2016regularization,laine2016temporal} demonstrate great promise in semi-supervised learning. The essential to this success relies on the key \emph{smoothness} assumption; that is, data points close to each other are likely to have the same label.
In our work, we inherit the spirit of these methods and design the regularization term as a consistency loss to encourage smooth predictions for the same data under different regularization or perturbations (\eg, Gaussian noise, network dropout, and randomized data transformation).
Then the regularization loss $R$ can be described as: 
\begin{equation}
R(\theta, \mathcal{L}, \mathcal{U}) =\sum_{i=1}^{N} \mathbb{E}_{\xi^{'}, \xi} \left \| f(x_i; \theta, \xi^{'}) - f(x_i;\theta, \xi) \right \| ^2,
\end{equation}
where $\xi$ and $\xi'$ denote different regularization or perturbations of input data.
In the following subsection, we will introduce how to effectively design the randomized data transformation regularization for the segmentation problem.


\subsection{Transformation Consistent Self-ensembling Model}
\begin{figure}[!t]
	\centering
	\includegraphics[width=0.7\linewidth]{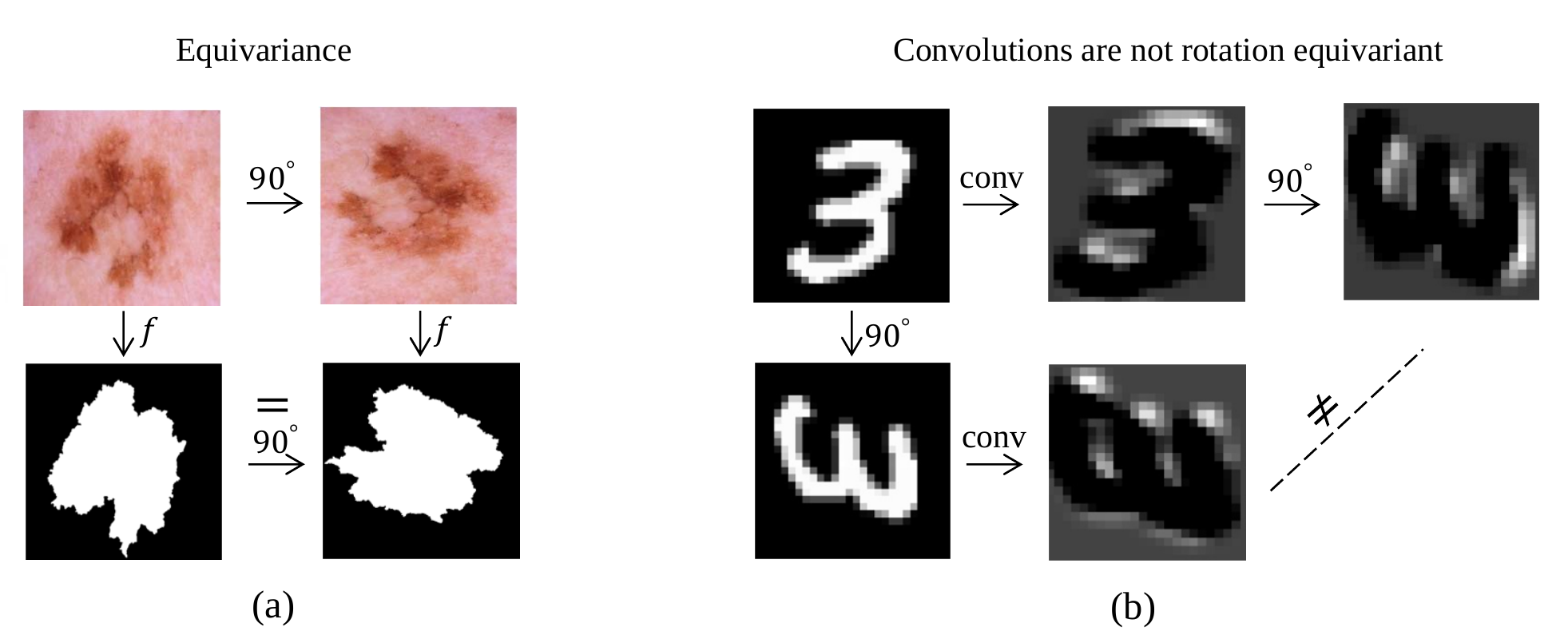}
	\caption{(a) Segmentation is desired to be rotation equivariant. If we rotate the input image, the excepted segmentation mask should have same rotation as the original segmentation mask. (b) Convolutions are not rotation equivariant in general.}
	\label{fig:method2}\centering
\end{figure}
Most regularization and pertrbations are easily designed for classification problem,
while we are confronted with a more challenging and practical skin lesion segmentation problem. 
One important difference is that the classification problem is transformation \emph{invariant} while the segmentation task is desired to be transformation \emph{equivariant}. 
Specifically, in the classification problem, we are only interested in the presence or absence of an object in the whole image, the classification result should remain the same, no matter what the data transformation (\ie, translation, rotation, and flipping) are applied to the input image. 
While in the segmentation task, if we rotate the input image, the expected segmentation mask should have the same rotation with original mask, although the corresponding pixel-wise predictions are same; see Figure~\ref{fig:method2}(a). 
However, in general, convolutions are not transformation (\ie, flipping, rotation) equivariant\footnote{Convolution is translation equivariant, and we focus on flipping and rotation transformation in this work.}, meaning that if one rotates or flips the CNN input, then the feature maps do not necessarily rotate in a meaningful or easy to predict manner~\cite{worrall2017harmonic}, as shown in Figure~\ref{fig:method2}(b).
Therefore, the convolutional network consisting of a series of convolutions is also not transformation equivariant. 
Formally, every transformation $ \pi \in \Pi$ of input $\textbf{x}$ associates with a transformation $\psi \in \Psi$  of the outputs; that is
$
\psi[f(\textbf{x})] = f(\pi[\textbf{x}]),
$
but in general $\pi \neq \psi$.

This property limits the unsupervised regularization effect of randomized data transformation for the segmentation problem~\cite{laine2016temporal}.
To enhance the regularization and more effectively utilize unlabeled data in our segmentation task, we introduce a transformation consistent scheme in the unsupervised regularization term. 
Specifically, this transformation consistent scheme is embedded into the framework by approximating $\psi$ to $\pi$ at the input and output space.
\begin{algorithm}[htb]  
	\caption{Model pseudocode.}  
	\label{alg:Framwork} 
	\small 
	\begin{algorithmic}[]  
		\Require $x_i \in {\mathcal{L}+\mathcal{U}} , y_i\in \mathcal{L}$
		\State $\lambda(t)$ =  unsupervised weight function
		\State
		$f_{\theta}(x)$ = neural network with trainable parameters $\theta$
		\State
		$\pi_i(x)$ =  transformation operation
		\For {$t$ in $[1, numepochs]$}
		\For {each minibatch $B$}
		\State randomly update $\pi_i(x)$
		\State  $z_{i \in B}$ $\leftarrow \pi_i(f_{\theta}(x_{i\in B})) $
		\State  $\tilde{z}_{i \in B}$ $\leftarrow f_{\theta}(\pi_i(x_{i\in B})) $
		\State $loss \leftarrow - \frac{1}{\left | B\cap \mathcal{L} \right |}\sum_{i\in(B\cap \mathcal{L})} {\rm log} z_i[\pi_i(y_i)]  + $ 
		\State  $\lambda(t)\frac{1}{\left | B \right |}\sum_{i\in B} 
		\left \| z_i - \tilde{z}_i \right \|^ 2$
		\State  update $\theta$ using optimizer
		\EndFor {}
		\EndFor 
		\State
		\Return $\theta$;  
	\end{algorithmic}  
	\label{alg1}
\end{algorithm}  
The pipeline of our proposed transformation consistent self-ensembling model is shown in Figure~\ref{fig:pipeline}, and the pseudocode is in Algorithm~\ref{alg1}. 
Each input $x_i$ is fed into the network for twice evaluation under transformation consistent scheme and other different perturbations (\eg, Gaussian noise and network dropout) to acquire two outputs $z_i$ and $\tilde{z}_i$.
Specifically, the transformation consistent scheme consists of triple $\pi_i$ operations; see Figure~\ref{fig:pipeline}. 
For one same training sample $x_i$, in the first evaluation, the operation $\pi_i$ is applied to the input image while in the second evaluation, the operation $\pi_i$ is applied on the prediction map. 
Through minimizing the difference between $z_i$ and $\tilde{z}_i$ with a mean square error loss function, we can regularize the network to be transformation consistent and further increase the network generalization capacity. 
Note that this regularization loss is applied for both labeled and unlabeled inputs.
For those labeled inputs $x_i \in \mathcal{L}$, we also apply the same operation $\pi_i$ to $y_i$ and use the standard cross-entropy loss to evaluate the correctness of network output.
Finally, the network is trained by minimizing the weighted combination of unsupervised regularization loss and supervised cross-entropy loss.
Note that we employed the same data augmentation in the training procedure of all the experiments for fair comparison. 
However, our method is different from traditional data augmentation.
Specifically, our method utilized the unlabeled data by minimizing network output difference under the transformed inputs, while obeying the \emph{smoothness} assumption.

\subsection{Training and Inference Procedures}
In the above transformation consistent scheme, we apply four kinds of rotation operations to the input with angles of $\gamma \cdot 90^{\circ}, $ where $ \gamma\in \left \{0,1,2,3 \right \}$.
We also apply a horizontal flipping operation. 
In total, eight possible transformation operations are obtained, and we randomly choose one operation in each training pass. 
We avoid the other angles for implementation simplification, but the proposed framework can be generalized to other angles in the future work. 
We employ the 2D DenseUNet-167 architecture in~\cite{li2018h} as our network backbone.
The dropout layer is applied after each convolutional layer in the encoding and decoding parts excepting for the last convolutional layer.
We use the standard data augmentation techniques on-the-fly to avoid overfitting. 
The data augmentation includes randomly flipping, rotating as well as scaling with a random scale factor from 0.9 to 1.1.
Note that all the experiments employed data augmentation for fair comparison.
The model was implemented using \emph{Keras} package~\cite{chollet2015keras}, and was trained with stochastic gradient descent (SGD) algorithm (momentum is 0.9 and minibatch size is 10).
The initial learning rate was 0.01 and decayed according to the equation $lr = lr * (1- iterations/total\_iterations)^ {0.9}$. 
In the inference phase, we remove the transformation operations in the network and do one single test with original input for fair comparison. 
After getting the probability map from the network, we first apply thresholding with 0.5 to get the binary segmentation result, and then use morphology operation, \ie, filling holes to get the final skin lesion segmentation result.

\section{Experiments and Results}
\subsection{Dataset and Evaluation Metrics}
We evaluate our method on the dataset of 2017 ISIC skin lesion segmentation challenge~\cite{codella2017skin}, which includes a training set with 2000 annotated dermoscopic images, a validation set with 150 images, and a testing set with 600 images.
Five evaluation metrics are calculated in the challenge to evaluate the segmentation performance, including Jaccard index (JA), dice coefficient (DI), pixel-wise accuracy (AC), sensitivity (SE) and specificity (SP). 
Note that the final rank is determined according to JA in the 2017 ISIC skin lesion segmentation challenge.

\begin{figure}[!t]
	\centering
	\includegraphics[width=0.65\linewidth]{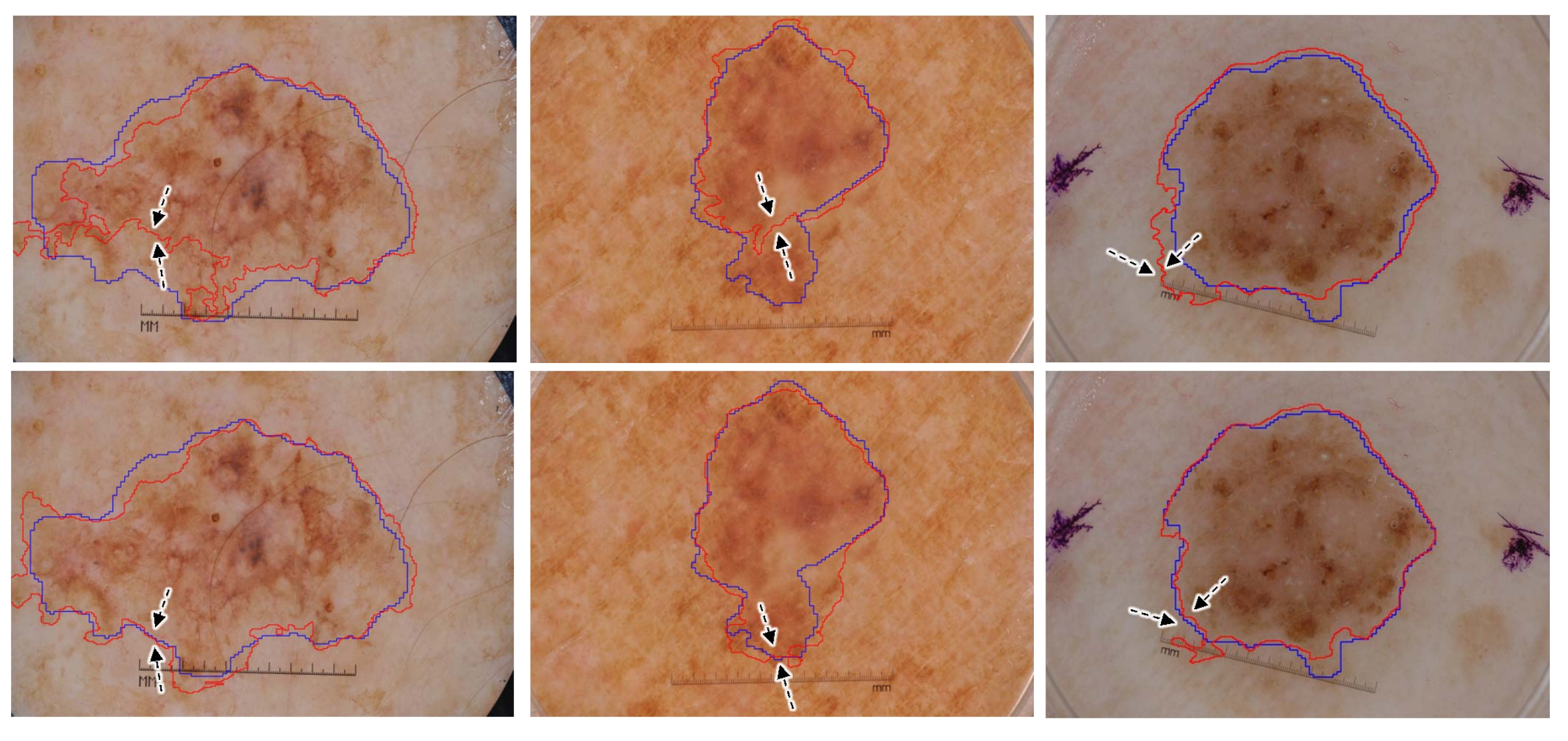}
	\caption{
		Segmentation results of supervised-only learning (top) and our method (bottom) on the validation dataset. The blue and red contours denote the ground truth and our result, respectively.}
	\label{fig:result}\centering
	
\end{figure}


\subsection{Analysis of Our Method}

\para{Quantitative and visual results with 50 labeled data.}\
In this part, we report the performance of our method trained with only 50 randomly selected labeled images and 1950 unlabeled images.
Table~\ref{tab:ablation} shows the experiments run with supervised-only method (the first one), supervised with regularization (the second one) and our semi-supervised method (the third one) on the validation dataset. 
The supervised-only experiment is trained with the same network backbone, but only optimized by the standard cross-entropy loss on the 50 labeled images.
It is obvious that compared with the supervised-only method, our semi-supervised method can achieve higher performance on all the evaluation metrics, with 2.46\%, 2.64\%, and 3.60\% improvements on JA, DI and SE, respectively. 
These prominent improvements on JA, DI and SE indicate that in most cases the false negative regions shrink while true positive regions expand to fit the true boundary of the lesion.
Comparing with the segmentation ground truth (blue contour), we can see the semi-supervised method can expand the segmented region to fit the ground truth lesion; see the left two examples in Figure~\ref{fig:result}.
Notably, our method would not simply amplify the segmentation result in all cases, it would also reasonably shrink the segmentations; see the examples in Figure~\ref{fig:result}.
These observations manifest that our semi-supervised learning method can improve the network generalization capability, compared with the supervised-only method.
It is worth mentioning that our method can also improve the supervised training; see Supervised with regularization in Table~\ref{tab:ablation}.


\begin{table}[ht]
	\begin{varwidth}[b]{0.5\linewidth}
		\centering
		\resizebox{1.05\columnwidth}{!}{
			\begin{tabular}{c|c|c|c}
				\hline
				\centering
				\multirow{2}{*}{Model} & \multicolumn{3}{C{3cm}}{50 labeled, 1950 unlabeled data}
				\tabularnewline
				\cline{2-4}
				& JA   & DI & SE \tabularnewline
				\hline
				Supervised-only & 72.85 &  81.15  & 82.77 \tabularnewline
				\hline
				Supervised with regularization & 73.25 &  81.60 & 83.30 \tabularnewline
				\hline
				Our Method & \textbf{75.31} & \textbf{83.79} & \textbf{86.37} \tabularnewline
				\hline
				\hline
				Our Method-A & 74.59 &  83.27 & 82.77 \tabularnewline
				\hline
				Our Method-B & 74.21 &  82.68 & 83.15 \tabularnewline
				\hline
			\end{tabular}}
		\caption{Comparison of supervised learning and semi-supervised learning on the validation dataset.}
		\label{tab:ablation}
	\end{varwidth}%
	\hfill
	\begin{minipage}[b]{0.45\linewidth}
		\centering
		\includegraphics[width=50mm]{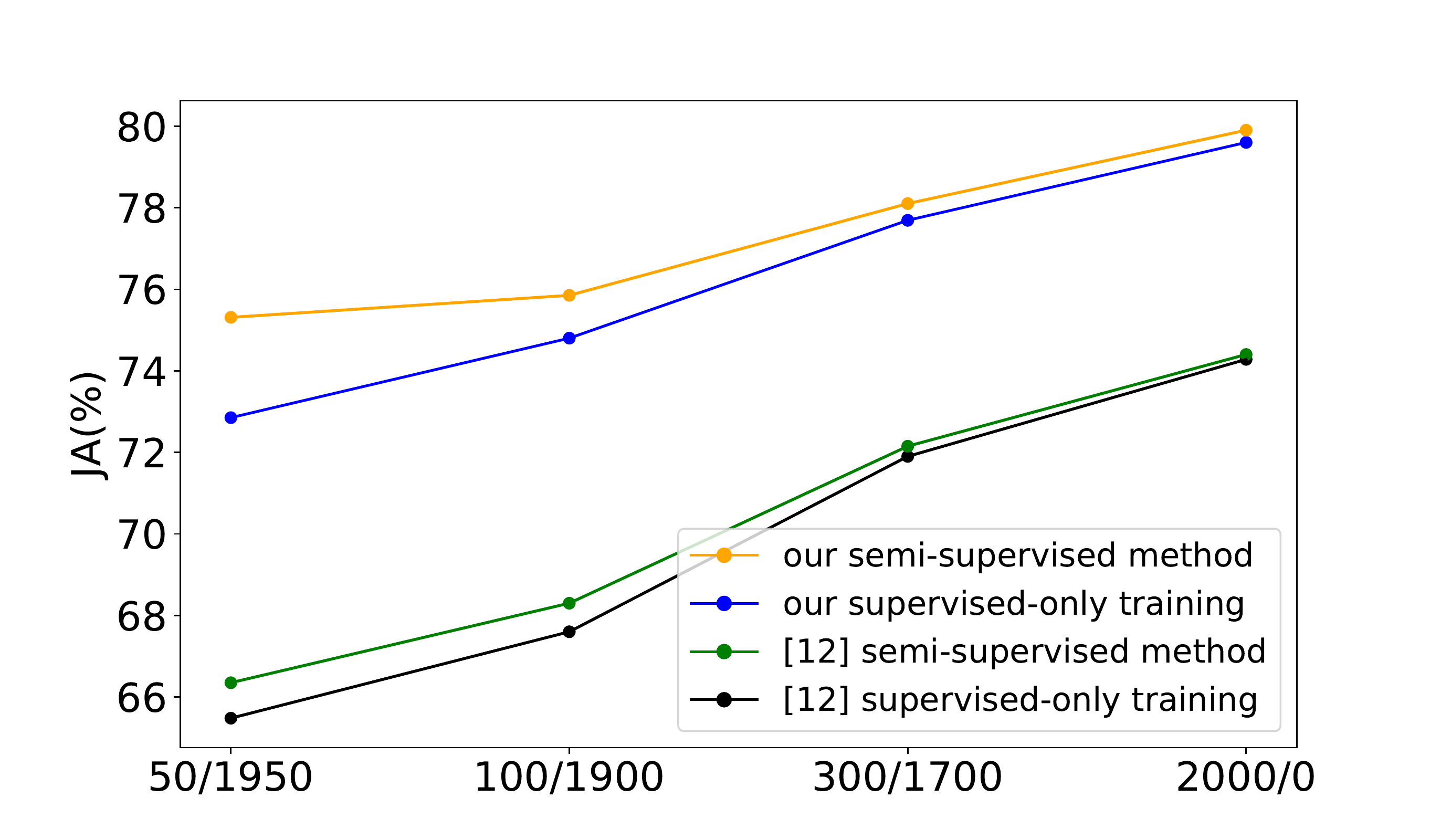}
		\captionof{figure}{Results with different number of labeled/unlabeled data.} 
		\label{fig:difnumber}
	\end{minipage}
\end{table}

\para{Effectiveness of transformation consistent scheme.}\
To show the effectiveness of the transformation consistent regularization scheme, we perform ablation analysis on our method.
We compare our method with the most common perturbations regularization, i.e., Gaussian noise and network dropout.
Table~\ref{tab:ablation} shows the experiment results, where ``Our Method-A" refers to semi-supervised learning with Gaussian noise and dropout regularization, ``Our Method-B" denotes to semi-supervised learning with transformation consistent regularization, and ``Our Method" refers to the experiment with all of these regularizations.
Note that these experiments are performed on the same training data with 50 labels.
As shown in Table~\ref{tab:ablation}, both kinds of regularizations can independently contribute to the performance gains in semi-supervised learning.
The performance improvement with transformation consistent regularization is very competitive, compared with the performance increment with Gaussian noise and dropout regularizations.
We also see that these two regularizations are complementary.
When the two kinds of regularizations are employed, the performance can be further boosted.

\para{Results under different number of labeled data.}\
Figure~\ref{fig:difnumber} demonstrates the network performance under different number of labeled images.
We draw the JA score of our semi-supervised method (trained with labeled data and unlabeled data) and supervised-only training (trained only with labeled data).
We can observe that the semi-supervised method consistently performs better than supervised-only in different labeled/unlabeled data settings, which demonstrates that our method effectively utilizes unlabeled data and is beneficial to the performance gains.
Note that in all semi-supervised learning experiments, we train the network with 2000 images in total, including labeled images and unlabeled images.
As expected, the performance of supervised-only training increases when more labeled training images are available; see the blue line in Figure~\ref{fig:difnumber}.
At the same time, the segmentation performance of semi-supervised learning can also be increased with more labeled training images; see the orange line in Figure~\ref{fig:difnumber}.
However, as we add more labeled samples, the difference in segmentation accuracy between semi-supervised and supervised-only becomes smaller.
The observation conforms with our expectation that our method leverages the distribution information from the unlabeled dataset.
As the labeled dataset is small, our method can gain a large improvement, since the regularization loss can effectively leverage the data distribution information from the unlabeled data.
Comparatively, as the labeled data increases, the improvements become limited. 
This is because both labeled and unlabeled images are randomly selected from the same dataset. When we have more labeled images, our regularization term can benefit from  less additional distribution information from the unlabeled data.
In the clinical practice, our approach is highly promising when a large number of unlabeled data from different protocols are acquired every day.

\para{Robustness analysis.}\
As we mentioned above, our semi-supervised method can improve the robustness of the network due to the regularization effect of the unsupervised loss.
From the comparison between the semi-supervised method and supervised method trained with 2000 labeled images in Figure~\ref{fig:difnumber}, we can see that our method can increase the JA performance when all labels are used (from 79.60\% to 80.02\%).
Note that the unsupervised loss was employed on all the input data and both experiments used the same data augmentation. 
Therefore, 
the improvement indicates that the unsupervised loss can provide a strong regularization to the labeled data, which would be useful in the case that the ground truth is not accurate due to the ambiguous lesions; see Figure~\ref{fig:intro}.   
In other words, the consistency requirement in the regularization term can encourage the network to learn more robust features and has a potential to be tolerant to ambiguous labels. 

\begin{table*}[!t]
	\centering
	\caption{Comparison on the test dataset in the ISIC 2017 skin lesion segmentation challenge.}
	\label{tab: testdata} %
	{	\resizebox{0.65\columnwidth}{!}{
			\begin{tabular}{c|c|c|c|c|c}
				\toprule[1.5pt]
				Team & JA  & DI  & AC & SE & SP \tabularnewline  \hline \hline
				Our Semi-supervised Method
				& \textbf{0.798} &  \textbf{0.874}  &  \textbf{0.943}  & \textbf{0.879} & 0.953 \tabularnewline
				\hline
				
				Our Baseline	& 0.772 &  0.853  &  0.936  & 0.837 & 0.969 \tabularnewline
				\hline
				~\citet{yuan2017improving} & 0.765 & 0.849   & 0.934  & 0.825 & 0.975  \tabularnewline
				\hline
				~\citet{berseth2017isic} & 0.762 & 0.847 &  0.932 & 0.820 & 0.978  \tabularnewline
				\hline
				~\citet{bi2017automatic} & 0.760 & 0.844 &  0.934  & 0.802 & \textbf{0.985}  \tabularnewline
				\hline
				RECOD & 0.754 & 0.839 & 0.931 & 0.817 & 0.970  \tabularnewline
				\hline		
				Jer & 0.752 & 0.837 &  0.930 & 0.813 & 0.976  \\  \bottomrule[1.5pt]	  
			\end{tabular}}

		}
	\end{table*}

\subsection{Comparison with Other Methods}

\begin{table*}[!t]
	\centering
	\caption{JA performance of different semi-supervised methods on the validation dataset.}
	
	\label{tab: semi_comparision} %
	{	\resizebox{0.65\columnwidth}{!}{
			\begin{tabular}{c|c|c|c}
				\toprule[1.5pt]
				Method 								& 50 labeled  & 50 labeled and 1950 unlabeled  &  Improvement \tabularnewline \hline 
				Our 								& 0.7285  &	0.7531	& 0.0246 \tabularnewline \hline
				Bai \etal~\cite{bai2017semi}		& 0.7285  & 0.7440  & 0.0155 \tabularnewline \hline
				Huang \etal~\cite{Hung_semiseg_2018}& 0.6548  & 0.6635	& 0.0087 \\ \bottomrule[1.5pt]	  
				
			\end{tabular}}
	}
	\vspace{-0.3cm}
\end{table*}
We compare our method with state-of-the-art methods submitted to the ISIC 2017 skin lesion segmentation challenge.
There are totally 21 submissions and the top results are listed in Table~\ref{tab: testdata}.
We trained two models: semi-supervised learning with 300 labeled data and 1700 unlabeled data and supervised-only network with 300 labeled data. 
We refer the last experiment as our baseline model. 
As shown in Table~\ref{tab: testdata}, our semi-supervised method achieved the best performance on the benchmark, outperforming the state-of-the-art method~\cite{yuan2017improving} with 3.3\% improvement on JA (from 76.5\% to 79.8\%). 
The performance gains on DI and SE are consistent with that on JA, with 2.5\% and 5.4\% improvement, respectively.
Our baseline model with 300 labeled data also excels the other methods due to the state-of-the-art network architecture.
Based on this architecture, our semi-supervised learning method further makes significant improvements, which demonstrate the effectiveness of the overall semi-supervised learning method.

We also compare our method with the latest semi-supervised segmentation method~\cite{bai2017semi} in the medical imaging community and an adversarial learning based semi-supervised method
\cite{Hung_semiseg_2018}. 
We conduct experiments with the setting of 50 labeled images and 1950 unlabeled images.
Table~\ref{tab: semi_comparision} shows the JA performance of different methods.
As shown in Table~\ref{tab: semi_comparision}, our proposed method achieves 2.46\% JA improvement by utilizing unlabeled data. However, ~\cite{bai2017semi} and ~\cite{Hung_semiseg_2018} can only enhance 1.55\% and 0.87\% improvement on JA, respectively.
Due to the different network backbone (DenseUNet-167 in~\cite{bai2017semi} and FC-ResNet101 in~\cite{Hung_semiseg_2018}), the performance with 50 labeled data is different. 
Figure~\ref{fig:difnumber} also shows the performance improvement of semi-supervised learning scheme of our method and \cite{Hung_semiseg_2018} under the setting of 100, 300 and 2000 labeled data.
We can see that the improvement of~\cite{Hung_semiseg_2018} is inferior than our method in all labeled/unlabeled settings, which also validates the effectiveness of our method.

\section{Conclusion}
In this paper, we present a novel semi-supervised learning method for skin lesion segmentation.
Specifically, we introduce a novel transformation consistent self-ensembling model for the segmentation task, which enhances the regularization effects to utilize the unlabeled data.
Comprehensive experimental analysis on the ISIC 2017 skin lesion segmentation challenge dataset demonstrates the effectiveness of our semi-supervised learning and the robustness of our method. 
Our method is general enough and can be extended to other semi-supervised learning problems in medical imaging field.
In the future, we will explore more regularization forms and ensembling techniques for better leveraging of the unlabeled data.

\section{Acknowledgments}
We thank anonymous reviewers for the comments and suggestions.  The work is supported by the Research Grants Council of the Hong Kong
Special Administrative Region (Project no. GRF 14203115).
\bibliographystyle{bmvc2k}
\bibliography{refs}

\end{document}